\documentclass[sigconf]{acmart}
\usepackage{enumitem}

\copyrightyear{2026}
\acmYear{2026}
\setcopyright{cc}
\setcctype{by}
\acmConference[SIGIR '26]{Proceedings of the 49th International ACM SIGIR Conference on Research and Development in Information Retrieval}{July 20--24, 2026}{Melbourne, VIC, Australia}
\acmBooktitle{Proceedings of the 49th International ACM SIGIR Conference on Research and Development in Information Retrieval (SIGIR '26), July 20--24, 2026, Melbourne, VIC, Australia}
\acmDOI{10.1145/3805712.3809947}
\acmISBN{979-8-4007-2599-9/2026/07}

\AtBeginDocument{%
  }

\usepackage{booktabs}      
\usepackage{makecell}      
\usepackage{adjustbox}     
\usepackage{array}         
\usepackage{float}
\usepackage{multirow}
\usepackage{algorithm}
\usepackage{algorithmic}

\begin{document}

\title{MEG-RAG: Quantifying Multi-modal Evidence Grounding for Evidence Selection in RAG}

\author{Xihang Wang}
\authornote{These authors contributed equally.}
\affiliation{%
  \institution{Zhejiang University}
  \city{Hangzhou}
  \country{China}
}
\email{xihangwang@163.com}

\author{Zihan Wang}
\authornotemark[1]
\affiliation{%
  \institution{Peking University}
  \city{Beijing}
  \country{China}}
\email{1104976867@qq.com}


\author{Chengkai Huang}
\authornote{Corresponding author.}
\affiliation{%
  \institution{University of New South Wales,}
  \institution{Macquarie University} \city{Sydney}
 \country{Australia}}
 \email{chengkai.huang1@unsw.edu.au}

\author{Quan Z. Sheng}
\affiliation{%
  \institution{Macquarie University}
    \city{Sydney}
  \country{Australia}}
\email{michael.sheng@mq.edu.au}

\author{Lina Yao}
\affiliation{%
  \institution{University of New South Wales,}
  \institution{CSIRO’s Data61}
    \city{Sydney}
  \country{Australia}}
\email{lina.yao@unsw.edu.au}

\renewcommand{\shortauthors}{Xihang et al.}



\begin{abstract}
Multimodal Retrieval-Augmented Generation (MRAG) addresses key limitations of Multimodal Large Language Models (MLLMs), such as hallucination and outdated knowledge. However, current MRAG systems struggle to distinguish whether retrieved multimodal data truly supports the semantic core of an answer or merely provides superficial relevance. Existing metrics often rely on heuristic position-based confidence, which fails to capture the informational density of multimodal entities. To address this, we propose \textbf{Multi-modal Evidence Grounding (MEG)}, a semantic-aware metric that quantifies the contribution of retrieved evidence. Unlike standard confidence measures, MEG utilizes \textit{Semantic Certainty Anchoring}, which dynamically filters out high-frequency stopwords via Inverse Document Frequency (IDF) to focus strictly on information-bearing tokens. Building on MEG, we introduce \textbf{MEG-RAG}, a framework that trains a multimodal reranker to align retrieved evidence with the semantic anchors of the ground truth. By prioritizing high-value content based on semantic grounding rather than token probability distributions, MEG-RAG improves the accuracy and multimodal consistency of generated outputs. Extensive experiments on the $M^2$RAG benchmark show that MEG-RAG consistently outperforms strong baselines and demonstrates robust generalization across different teacher models.
The data and code are available at \href{https://anonymous.4open.science/r/anonym-9QM02BD/}{\underline{here}}.
\end{abstract}



\maketitle

\section{Introduction}


In recent years, multimodal large language models (MLLMs) have achieved rapid progress, demonstrating remarkable capabilities in understanding and generating multimodal content such as text, images, and audio \cite{jiao2026prunerag,ye2026memweaver,mei2025survey,huang2025towards,lou2025speechagent}. However, MLLMs still suffer from hallucination, outdated knowledge, and insufficient factual grounding \cite{wang2026scenealign,huang2025embedding}. To alleviate these issues, the Multimodal Retrieval-Augmented Generation (MRAG) paradigm has emerged, which retrieves external multimodal evidence (e.g., text snippets, images, audio clips) to enhance the factuality and contextual grounding of MLLMs' outputs. 

Despite the promise of MRAG, its effectiveness heavily depends on the quality of the retrieved documents. Recent research highlights that irrelevant or contradictory information can mislead the model and degrade generation quality \cite{mortaheb2025re,chen2024mllm,jiao2026doctor}. However, existing multimodal retrievers and rerankers are ill-equipped to filter out such documents, as they mainly rely on semantic similarity rather than quantifying the information's true contribution to accurate answer generation. Several works have explored multimodal rerankers to refine document selection \cite{chen2024mllm,xu2025mm}, but these methods primarily focus on improving relevance rather than measuring contribution. Although benchmarks such as M$^2$RAG \cite{liu2025benchmarking} have been proposed to standardize multimodal retrieval evaluation, they still lack a principled mechanism to quantify the real informational value of retrieved multimodal data \cite{mei2025survey,abootorabi2025ask}.

To address this limitation, we propose \textbf{MEG-RAG}, a framework for quantifying \textbf{Multi-modal Evidence Grounding (MEG)} and guiding evidence selection in MRAG. 
Specifically, MEG measures the contribution of each retrieved multimodal sample by evaluating the change in an MLLM’s semantic confidence with and without that sample. 
We then train a multimodal reranker to predict MEG scores, enabling effective re-ranking of multimodal evidence. 
This reranker can be seamlessly integrated into various MRAG architectures.

In this paper, our main contributions are threefold:

\begin{itemize}
    \item We propose the \textit{Multi-modal Evidence Grounding (MEG)}, which quantifies the contribution of each retrieved multimodal sample by measuring the change in an MLLM’s semantic confidence with and without that sample.
    \item Based on MEG, we develop \textit{MEG-RAG}, which leverages a multi-task reranker that jointly optimizes binary discrimination and pairwise ranking objectives, enabling precise selection of high-value multimodal evidence.
    \item Experiments on the M$^2$RAG benchmark, including \textit{open-domain question answering}, \textit{image captioning}, and \textit{fact verification}, show that MEG-RAG consistently outperforms strong baselines such as Jina Reranker, while achieving strong cross-task generalization.
\end{itemize}

\begin{figure*}[ht]
    \centering
    \includegraphics[width=\linewidth]{img/image_v11.png}
    \caption{Overview of the MEG-RAG framework. \textbf{(1) Dataset Construction} computes MEG scores to quantify the utility of multimodal documents. \textbf{(2) Reranker Training} learns a multimodal reranker using the constructed MEG dataset with a hybrid objective combining document relevance classification and pairwise ranking. \textbf{(3) MEG-RAG Inference} retrieves candidate documents, reranks them to select the top-$K$ results, and generates the final answer based on the selected evidence.}
    \label{fig:method_overview}
\end{figure*}

\section{Methodology}

In this section, we introduce the MEG-RAG to address the unique challenges of evaluating multimodal documents by introducing a new metric, Multi-modal Evidence Grounding (MEG). We then detail a multi-task optimization strategy to train a reranker that can effectively rank the multimodal documents through MEG prediction.

\subsection{Task Definition}

We consider the task of MRAG, where a model generates an answer $\mathrm{g}$ for a query $\mathrm{x}$ by leveraging a large-scale multimodal corpus $\mathcal{C}$. State-of-the-art MRAG frameworks employ a two-stage \textit{retrieve-then-rerank} pipeline to select the most relevant context. The end-to-end process can be formally expressed as:
{
\small
\begin{align}
    \mathcal{D}_{\text{cand}} &= \mathrm{Retriever}(\mathrm{x}, \mathcal{C}), \label{eq:retrieval} \\
    \mathcal{D}_{\text{final}} &= \mathrm{Reranker}(\mathrm{x}, \mathcal{D}_{\text{cand}}), \label{eq:rerank} \\
    \mathrm{g} &= \mathrm{MLLM}(\mathrm{x}, \mathcal{D}_{\text{final}}),
    \label{eq:generation}
\end{align}
}
First, an efficient initial retriever recalls a broad candidate set of documents $\mathcal{D}_{\text{cand}}$. Subsequently, a reranker model refines this set by performing a fine-grained relevance assessment. It selects a small, high-precision subset of documents $\mathcal{D}_{\text{final}}$. This refined context $\mathcal{D}_{\text{final}}$ is then prepended to the query $\mathrm{x}$ and fed into a Multimodal Large Language Model (MLLM) to produce the final answer $\mathrm{g}$.

\subsection{Multi-modal Evidence Grounding (MEG)}
	
To train a high-performance reranker, we require a robust supervisory signal that accurately reflects the semantic utility of a retrieved document. To this end, we introduce a novel metric, \textbf{Multi-modal Evidence Grounding (MEG)}, designed to quantify the marginal contribution of a document to the factual grounding of a correct answer. MEG measures the change in a teacher MLLM's semantic confidence when a given document is added to the context. We present our automated MEG data collection pipeline below.

\subsubsection{MEG Definition}
The core of our framework is to quantify the grounding contribution of a retrieved \textit{multimodal document}. Formally, given an MLLM $\phi$, a multimodal query $\mathrm{x}$, and its corresponding ground truth textual answer $\mathrm{y}$, the Multi-modal Evidence Grounding (MEG) for a retrieved document $\mathrm{d}_i$ is defined as:
\begin{equation}\label{eq:MEG_def}
\footnotesize
\mathrm{MEG}(\mathrm{d}_i|\mathrm{x}) \overset{def}{=} S_\phi(\mathrm{y}|\mathrm{x},\mathrm{d}_i) - S_\phi(\mathrm{y}|\mathrm{x}),
\end{equation}
where $S_\phi(\mathrm{y}|\cdot)$ represents the MLLM's semantic confidence in generating the correct answer $\mathrm{y}$ based on its core informational content. By measuring the change in semantic confidence, MEG captures the marginal utility of each retrieved document and provides a rich, continuous supervisory signal. A high MEG score indicates that the document offers critical grounding for the entities, relations, or actions in the answer; a near-zero score suggests that the document is largely redundant with respect to the semantic core; and a negative score indicates that the document introduces distracting or contradictory information that reduces grounding confidence.

\subsubsection{Semantic Certainty Anchoring}

A key challenge is reliably estimating the MLLM's confidence $S_\phi(\mathrm{y}|\cdot)$ without being biased by syntactic filler tokens (e.g., "the", "is", "of") in the ground truth answer. We propose \textbf{Semantic Certainty Anchoring}. This mechanism dynamically identifies information-bearing tokens to serve as anchors for confidence calculation based on their statistical significance.

Specifically, for a ground truth answer $\mathrm{y} = (t_1, t_2, ..., t_L)$, we apply a semantic mask $\mathcal{M} \in \{0, 1\}^L$. Instead of relying on external linguistic tools, we employ a frequency-based filtering strategy. We calculate the inverse document frequency (IDF) for each word $t_i$ across the corpus. Words with high frequency (low IDF) are typically stopwords that contribute little to semantic distinction. The mask is defined as:
\begin{equation}
\mathcal{M}_i = 
\begin{cases} 
1 & \text{if } \text{IDF}(t_i) > \tau_{\text{freq}} \\
0 & \text{otherwise}
\end{cases}
\end{equation}
where $\tau_{\text{freq}}$ is a threshold determining the cut-off for high-frequency words. $\mathcal{M}_i$ is set to 1 for rare, information-rich tokens and 0 for common, low-information tokens. The semantic confidence is defined as the average log-probability over these anchors:
{
\small
\begin{equation}\label{eq:semantic_p_phi}
S_\phi(\mathrm{y}|\cdot) \overset{\text{def}}{=} \frac{1}{\sum_{i=1}^L \mathcal{M}_i} \sum_{i=1}^L \mathcal{M}_i \cdot \log p_\phi(t_i | t_{<i}, \cdot)
\end{equation}
}
This data-driven approach ensures that the supervision signal focuses on the semantic essence of the multimodal task rather than heuristic positional biases or linguistic priors. Notably, our MEG formulation relies on the difference of log-probabilities rather than pure probability subtraction or Kullback-Leibler (KL) divergence. From an information-theoretic perspective, this log-probability difference acts as an estimator of Pointwise Mutual Information (PMI) \cite{church1990word}, robustly capturing the information gain provided by the evidence while avoiding numerical underflow. Furthermore, recent empirical studies demonstrate that log-probabilities serve as a highly reliable metric for evaluating semantic plausibility and world knowledge in both base and instruction-tuned Large Language Models \cite{kauf2024log}. Finally, unlike KL divergence, which requires computing and storing the full vocabulary distribution at each generation step, our approach only requires the probability of the ground-truth semantic anchors, making it highly computationally efficient for MLLMs. We collect a rich training dataset characterized by the MEG of multimodal documents, providing a foundation for reranker training.

\subsection{Multi-task Reranker Optimization}
Building on the MEG-scored dataset, we train a multimodal reranker $f_\theta$ to predict document utility via a hybrid-loss scheme that incorporates both absolute relevance (classification) and relative utility (ranking) objectives. Our framework is model-agnostic and allows for the utilization of any Transformer-based multimodal encoder as the backbone. The reranker takes the query-document pair as input and outputs a scalar utility score.

\subsubsection{Document Relevance Classification}
To enable the reranker to learn the absolute relevance between the document and query, we frame the task as a binary classification problem. We employ a standard Cross-Entropy (CE) loss:
\begin{equation}
\footnotesize
\begin{aligned}
\min_{\theta} \quad L_{\text{CE}} = & \frac{1}{N} \sum_{i=1}^N \Big[-y_i \log(p_i) -(1-y_i)\log(1-p_i)\Big], \\
\end{aligned}
\label{eq:ce_loss}
\end{equation}

\noindent where $p_i = f_\theta(\mathrm{x}_i, \mathrm{d}_i)$ is the probability that document $\mathrm{d}_i$ is helpful for query $\mathrm{x}_i$. The binary label $y_i$ is determined using two thresholds, $b_1$ and $b_2$, on the MEG score of the document: documents with $\mathrm{MEG} > b_1$ are positive ($y_i=1$), while those with $\mathrm{MEG} < b_2$ are negative ($y_i=0$).

\subsubsection{Pairwise Ranking Optimization}
While CE loss assesses absolute relevance, effective RAG requires a nuanced relative ordering of documents. To achieve this, we incorporate a pairwise ranking objective based on RankNet loss \cite{burges2005learning}. For a given query $\mathrm{x}$, we sample pairs of positive $(\mathrm{d}_i)$ and negative $(\mathrm{d}_j)$ documents based on their MEG scores. The reranker $f_\theta$ produces scalar scores $s_i = f_\theta(\mathrm{x}, \mathrm{d}_i)$ and $s_j = f_\theta(\mathrm{x}, \mathrm{d}_j)$. The RankNet loss minimizes the cross-entropy of the "correct" ranking probability:

\begin{equation}
\footnotesize
\begin{aligned}
\min_{\theta} \quad L_{\text{RankNet}} = & \sum_{(\mathrm{x}, \mathrm{d}_i, \mathrm{d}_j) \in \mathcal{T}} \log\left(1 + \exp\left(-\sigma(s_i - s_j)\right)\right),
\end{aligned}
\label{eq:ranknet_loss}
\end{equation}

\noindent where $\mathcal{T}$ is the set of all training pairs $(\mathrm{d}_i, \mathrm{d}_j)$ where $\mathrm{MEG}(\mathrm{d}_i|\mathrm{x}) > \mathrm{MEG}(\mathrm{d}_j|\mathrm{x})$, and $\sigma$ is a scaling factor that controls the steepness of the logistic function. This loss directly encourages the reranker to assign a higher score ($s_i$) to the document with higher MEG.

\subsubsection{Final Training Objective}
The complete training objective for our multimodal reranker combines both the classification and ranking losses, balanced by a hyperparameter $\alpha$:
\begin{equation}
\footnotesize
L_{\text{hybrid}} = \alpha \cdot L_{\text{CE}} + (1-\alpha) \cdot L_{\text{RankNet}}.
\label{eq:total_loss}
\end{equation}
This dual capability is critical for selecting the optimal top-k multimodal evidence to pass to the MLLM, thereby improving the final answer's quality and factual grounding.

\begin{table*}[ht]
\small
\setlength{\tabcolsep}{3pt} 
\centering
\vspace{-1em}
\caption{Main comparison of MEG-RAG against baselines on the $M^2RAG$ benchmark tasks across multiple MLLMs. (B-N = Bleu-N, R-N = Rouge-N, R-L = Rouge-L). \textbf{Bold} indicates the best performance, and \underline{underline} indicates the second-best performance.}
\vspace{-1em}
\label{tab:main_results}
\begin{adjustbox}{max width=\textwidth}
\begin{tabular}{ll|cccccccc|cccccccc|cc}
\toprule[1.5pt]
\multirow{2}{*}{\textbf{Model}} & \multirow{2}{*}{\textbf{Method}} & \multicolumn{8}{c|}{\textbf{Multimodal QA}} & \multicolumn{8}{c|}{\textbf{Image Captioning}} & \multicolumn{2}{c}{\textbf{Fact Verification}} \\
& & \textbf{B-1} & \textbf{B-2} & \textbf{B-3} & \textbf{B-4} & \textbf{R-1} & \textbf{R-2} & \textbf{R-L} & \textbf{CIDEr} & \textbf{B-1} & \textbf{B-2} & \textbf{B-3} & \textbf{B-4} & \textbf{R-1} & \textbf{R-2} & \textbf{R-L} & \textbf{CIDEr} & \textbf{ACC} & \textbf{F1} \\
\midrule[1pt]
\multirow{7}{*}{\makecell{LLaVA-NeXT}} 
& None-MRAG & 34.85 & 30.00 & 27.04 & 24.49 & 48.36 & 37.00 & 45.61 & 250.82 & 6.42 & 3.08 & 2.02 & 1.63 & 19.17 & 5.34 & 17.33 & 18.98 & 39 & 35.76 \\
\cmidrule{2-20}
& Naive-MRAG-top1 & 35.73 & 30.77 & 27.51 & 24.86 & 49.17 & \underline{37.44} & 46.28 & 251.26 & 8.99 & 5.32 & 3.78 & \underline{2.82} & 22.17 & 8.43 & 20.01 & 40.34 & 50 & 33.18 \\
& Jina-Reranker-top1 & \underline{35.94} & \underline{30.97} & \underline{27.73} & \underline{25.10} & \underline{49.47} & 37.40 & \underline{46.97} & \underline{254.57} & \underline{9.55} & \underline{5.65} & \underline{3.91} & 2.80 & \underline{26.26} & \underline{9.94} & \underline{23.44} & \underline{42.08} & \underline{52} & \underline{34.93} \\
& \textbf{MEG-RAG-top1} & \textbf{37.28} & \textbf{32.41} & \textbf{29.24} & \textbf{26.69} & \textbf{51.41} & \textbf{40.11} & \textbf{49.19} & \textbf{268.74} & \textbf{12.73} & \textbf{8.45} & \textbf{6.33} & \textbf{4.63} & \textbf{30.01} & \textbf{13.73} & \textbf{26.75} & \textbf{65.59} & \textbf{58} & \textbf{41.13} \\
\cmidrule{2-20}
& Naive-MRAG-top3 & 35.73 & 30.81 & 27.65 & 24.90 & 49.81 & 37.83 & 46.93 & 250.52 & 9.30 & 5.52 & 3.85 & 2.84 & 23.99 & 9.74 & 21.87 & 35.48 & 47 & 31.14  \\
& Jina-Reranker-top3 & \underline{36.19} & \underline{31.37} & \underline{28.32} & \underline{25.78} & \underline{50.08} & \underline{38.37} & \underline{47.44} & \underline{259.41} & \underline{10.05} & \underline{5.98} & \underline{4.10} & \underline{3.02} & \underline{25.79} & \underline{10.68} & \underline{22.91} & \underline{44.60} & \underline{50} & \underline{34.26} \\
& \textbf{MEG-RAG-top3} & \textbf{37.89} & \textbf{33.16} & \textbf{30.21} & \textbf{27.71} & \textbf{52.66} & \textbf{41.01} & \textbf{50.22} & \textbf{278.43} & \textbf{13.17} & \textbf{8.18} & \textbf{5.95} & \textbf{4.43} & \textbf{29.85} & \textbf{12.84} & \textbf{25.96} & \textbf{57.78} & \textbf{55} & \textbf{38.25} \\
\midrule[1pt]
\multirow{7}{*}{\makecell{Qwen2.5-VL}} 
& None-MRAG & 25.41 & 20.64 & 17.61 & 14.91 & 39.30 & 27.78 & 34.97 & 105.33 & 8.97 & 4.46 & 2.84 & 2.03 & 20.05 & 6.64 & 18.20 & 19.51 & 51 & 50.67 \\
\cmidrule{2-20}
& Naive-MRAG-top1 & 30.04 & 25.15 & 22.16 & \underline{19.90} & 43.43 & 32.11 & 40.43 & \underline{183.07} & 11.63 & 6.10 & 3.58 & 2.53 & \underline{25.21} & 8.68 & 22.64 & 40.29 & 56 & 56.63  \\
& Jina-Reranker-top1 & \underline{30.76} & \underline{25.73} & \underline{22.51} & 19.70 & \underline{45.23} & \underline{33.56} & \underline{41.39} & 179.37 & \underline{11.90} & \underline{6.86} & \underline{4.41} & \underline{3.08} & 25.10 & \underline{9.39} & \underline{22.68} & \underline{43.27} & \underline{58} & \underline{59.23} \\
& \textbf{MEG-RAG-top1} & \textbf{32.50} & \textbf{27.49} & \textbf{24.40} & \textbf{21.85} & \textbf{47.85} & \textbf{35.22} & \textbf{44.17} & \textbf{211.24} & \textbf{14.71} & \textbf{8.21} & \textbf{5.68} & \textbf{4.12} & \textbf{29.19} & \textbf{13.86} & \textbf{26.62} & \textbf{50.00} & \textbf{62} & \textbf{63.74} \\
\cmidrule{2-20}
& Naive-MRAG-top3 & \underline{30.67} & \underline{25.45} & 22.14 & 19.55 & 45.04 & 32.63 & \underline{41.42} & \underline{185.94} & 11.75 & 6.92 & \underline{4.93} & \underline{3.70} & 26.99 & 10.77 & 24.55 & \underline{46.64} & 60 & 62.26 \\
& Jina-Reranker-top3 & 29.89 & 25.19 & \underline{22.16} & \underline{19.68} & \underline{45.41} & \underline{33.50} & 41.39 & 174.55 & \underline{12.62} & \underline{7.16} & 4.56 & 3.49 & \underline{27.82} & \underline{11.86} & \underline{25.43} & 45.73 & \underline{63} & \underline{65.96} \\
& \textbf{MEG-RAG-top3} & \textbf{33.37} & \textbf{28.11} & \textbf{24.79} & \textbf{22.02} & \textbf{47.94} & \textbf{36.03} & \textbf{44.73} & \textbf{218.62} & \textbf{15.31} & \textbf{9.77} & \textbf{7.10} & \textbf{5.58} & \textbf{31.24} & \textbf{16.29} & \textbf{28.50} & \textbf{73.66} & \textbf{70} & \textbf{73.75} \\
\midrule[1pt]
\multirow{7}{*}{\makecell{MM-RAIT- \\ Qwen2.5-VL}} 
& None-MRAG & 33.88 & 27.39 & 23.15 & 20.20 & 48.31 & 30.52 & 46.25 & 206.42 & 16.18 & 10.46 & 7.56 & 5.68 & 34.16 & 15.43 & 32.58 & 79.41 & 59 & 61.15 \\
\cmidrule{2-20}
& Naive-MRAG-top1 & 38.12 & 29.88 & 24.57 & 21.56 & 56.64 & 33.07 & 53.81 & 236.08 & 16.98 & 11.35 & 8.44 & 6.44 & 35.75 & 16.31 & 33.81 & 83.62 & 60 & 62.44 \\
& Jina-Reranker-top1 & \underline{39.16} & \underline{30.79} & \underline{25.33} & \underline{22.34} & \underline{58.29} & \underline{33.35} & \underline{55.70} & \underline{243.65} & \underline{18.66} & \underline{12.77} & \underline{9.57} & \underline{7.18} & \underline{35.84} & \underline{16.82} & \underline{33.81} & \underline{99.06} & \underline{62} & \underline{63.65} \\
& \textbf{MEG-RAG-top1} & \textbf{41.84} & \textbf{33.26} & \textbf{27.43} & \textbf{24.00} & \textbf{61.34} & \textbf{36.70} & \textbf{58.86} & \textbf{260.78} & \textbf{22.82} & \textbf{17.14} & \textbf{13.32} & \textbf{10.95} & \textbf{42.45} & \textbf{23.02} & \textbf{40.49} & \textbf{134.05} & \textbf{66} & \textbf{67.43}  \\
\cmidrule{2-20}
& Naive-MRAG-top3 & \underline{40.67} & 31.39 & 26.08 & 23.07 & \underline{56.56} & 33.37 & \underline{54.09} & 247.13 & 17.15 & 11.41 & 7.90 & 5.89 & 36.09 & 16.89 & 34.57 & 84.41 & 62 & 62.95 \\
& Jina-Reranker-top3 & 40.46 & \underline{31.61} & \underline{26.29} & \underline{23.41} & 56.34 & \underline{34.17} & 54.05 & \underline{248.63} & \underline{19.39} & \underline{13.05} & \underline{9.47} & \underline{7.42} & \underline{38.05} & \underline{18.31} & \underline{35.74} & \underline{96.66} & \underline{63} & \underline{64.31} \\
& \textbf{MEG-RAG-top3} & \textbf{42.95} & \textbf{33.38} & \textbf{27.60} & \textbf{24.14} & \textbf{61.24} & \textbf{35.89} & \textbf{57.91} & \textbf{264.78} & \textbf{23.47} & \textbf{17.46} & \textbf{13.74} & \textbf{11.36} & \textbf{42.91} & \textbf{24.10} & \textbf{41.24} & \textbf{136.44} & \textbf{68} & \textbf{69.80}\\
\bottomrule[1.5pt]
\end{tabular}
\end{adjustbox}
\end{table*}

\section{Experiments}

In this section, we conduct experiments to validate the effectiveness of our proposed MEG-RAG framework.

\textbf{Tasks and Datasets.} We evaluate MEG-RAG on three multimodal tasks from the M$^2$RAG benchmark \cite{liu2025benchmarking}: (1) \textbf{MultiModal QA} (MMQA), an open-domain task requiring the MLLM to synthesize information from retrieved multimodal documents to answer questions; (2) \textbf{Image Captioning}, where the MLLM generates a descriptive caption for an image, using retrieved documents as context; and (3) \textbf{Fact Verification}, which requires the model to verify the truthfulness of a multimodal claim based on retrieved evidence.

\textbf{Models and Metrics.} Our evaluations are conducted on both general MLLMs such as \textbf{LLaVA-NeXT} \cite{li2024llava}, \textbf{Qwen2.5-VL} \cite{bai2025qwen2} 
as well as \textbf{MM-RAIT-Qwen2.5-VL}, a variant of Qwen2.5-VL specifically fine-tuned for MRAG tasks \cite{liu2025benchmarking}. We follow the M$^2$RAG’s standard metrics: BLEU, ROUGE, and CIDEr for MMQA and Image Captioning tasks, and Acc and F1 for Fact Verification.

\textbf{Baselines.} We compare our method with three baselines: (1) None-MRAG, (2) Naive-MRAG, where top-$k$ retrieved documents are directly fed to the MLLM; (3) Ranking-MRAG, where a reranker is used to reorder retrieved documents such as Jina Reranker \cite{jina-m0-blog}.

\textbf{Implementation Details.} We constructed the training dataset based on the MMQA task. For the proposed Semantic Certainty Anchoring, we set the frequency threshold $\tau_{\text{freq}} = 0.15$ to exclude the redundant words in the ground truth answer, identifying the remaining words as semantic anchors. This threshold was empirically determined to maximize grounding quality on the validation set. Using the calculated MEG scores, we labeled $<$query, answer, document$>$ triplets. 

To ensure robust training and precise discrimination of evidence, we established a rigorous data filtering pipeline. An analysis of the initially generated raw triplets revealed a heavily skewed distribution: approximately $6.5\%$ were useful samples ($\mathrm{MEG} > 0.2$, providing strong semantic grounding), $12.9\%$ were misleading samples ($\mathrm{MEG} < -0.2$, causing significant interference), and the vast majority, $80.6\%$, were irrelevant or neutral samples ($-0.2 \le \mathrm{MEG} \le 0.2$). To prevent the model from being overwhelmed by noise and biased toward neutral contexts, all irrelevant samples falling within the $[-0.2, 0.2]$ interval were directly discarded. Finally, to address the imbalance between the remaining valid positive and negative pairs, we sampled exactly 31K positive triplets and 31K negative triplets. This yielded a strictly balanced final training dataset containing 62K triplets (a perfect 1:1 ratio). 

We instantiated our reranker using the Jina-m0 architecture \cite{jina-m0-blog}. The model was fine-tuned on four NVIDIA A100 (80GB) GPUs. We used the AdamW optimizer with a learning rate of 2e-5 and a batch size of 64 per GPU. The multi-task objective (\ref{eq:total_loss}) was optimized with the trade-off parameter $\alpha=0.74$. The training was conducted for 3 epochs with a linear warmup of the first 10\% steps. During inference, the reranker selects the top-$k$ ($k \in \{1, 3\}$) documents for generation.

\subsection{Main Result}

As shown in Table \ref{tab:main_results}, MEG-RAG consistently delivers significant improvements across a wide range of tasks and MLLMs. On the MM-RAIT-Qwen2.5-VL, our method achieves a top CIDEr score of 260.78 on the MMQA task under the top-1 setting, representing a relative improvement of 7.0\% over the Jina-reranker baseline (243.65). For the general MLLM LLaVA-NeXT, MEG-RAG attains a CIDEr score of 278.43 with the top-3 setting, marking a substantial 7.3\% gain compared to the Jina-reranker (259.41).
Furthermore, MEG-RAG delivers consistent gains on Image Captioning and Fact Verification. For Image Captioning, MEG-RAG (top-3) with Qwen2.5-VL achieves a CIDEr score of 73.66, substantially outperforming Jina Reranker (45.73) with a relative improvement of \textbf{61.1\%}. On Fact Verification, it attains an F1 of 73.75, surpassing the baseline (65.96) by \textbf{11.8\%}. Crucially, since our reranker is trained solely on the MMQA dataset, these cross-task gains strongly validate its robust generalization capability.

\subsection{Ablation Study}

To validate the effectiveness of our design, we conduct two key ablation studies.

\begin{table}[ht]
\small
\setlength{\tabcolsep}{4pt}
\centering
\caption{Ablation study of loss functions (top-3).} 
\vspace{-1em}
\label{tab:ablation_loss}
\begin{adjustbox}{max width=\columnwidth}
\begin{tabular}{ll|ccc|ccc|cc}
\toprule[1.5pt]
\multirow{2}{*}{\textbf{Model}} & \multirow{2}{*}{\textbf{Method (Loss)}} & \multicolumn{3}{c|}{\textbf{Multimodal QA}} & \multicolumn{3}{c|}{\textbf{Image Captioning}} & \multicolumn{2}{c}{\textbf{Fact Verification}} \\
& & \textbf{B-4} & \textbf{R-L} & \textbf{CIDEr} & \textbf{B-4} & \textbf{R-L} & \textbf{CIDEr} & \textbf{ACC} & \textbf{F1} \\
\midrule[1pt]
\multirow{3}{*}{\makecell{LLaVA-NeXT}} 
& CE Loss & 26.45 & 49.43 & 266.80 & 3.72 & 24.21 & 50.92 & 52 & 34.87 \\
& RankNet Loss & 26.33 & 49.36 & 266.61 & 3.41 & 23.68 & 45.41 & 51 & 34.52 \\
& \textbf{Hybrid-Loss} & \textbf{27.71} & \textbf{50.22} & \textbf{278.43} & \textbf{4.43} & \textbf{25.96} & \textbf{57.78} & \textbf{55} & \textbf{38.25} \\
\midrule[1pt]
\multirow{3}{*}{\makecell{MM-RAIT- \\ Qwen2.5-VL}} 
& CE Loss & 23.37 & 56.78 & 257.34 & 9.74 & 37.59 & 122.12 & 65 & 67.41 \\
& RankNet Loss & 23.00 & 56.04 & 254.93 & 8.53 & 38.04 & 106.55 & 64 & 65.01 \\
& \textbf{Hybrid-Loss} & \textbf{24.14} & \textbf{57.91} & \textbf{264.78} & \textbf{11.36} & \textbf{41.24} & \textbf{136.44} & \textbf{68} & \textbf{69.80} \\
\bottomrule[1.5pt]
\end{tabular}
\end{adjustbox}
\end{table}

\textbf{Impact of Loss Function.} We compare our hybrid-loss model against the single-loss model in Table \ref{tab:ablation_loss}. The Hybrid-Loss model consistently outperforms the single-loss variants across all MLLMs and tasks. For \textbf{LLaVA-NeXT} on the MMQA task, the Hybrid-Loss method yielded a BLEU-4 of \textbf{27.71}, which exceeds the RankNet loss (26.33) by $5.2\%$ and the CE loss (26.45) by $4.8\%$. 
This confirms that the two loss components capture complementary learning signals, namely absolute relevance (CE) and relative ordering (RankNet), and that their synergistic combination is essential for achieving optimal reranker performance.

\begin{figure}[t]
    \centering
    \includegraphics[width=\linewidth]{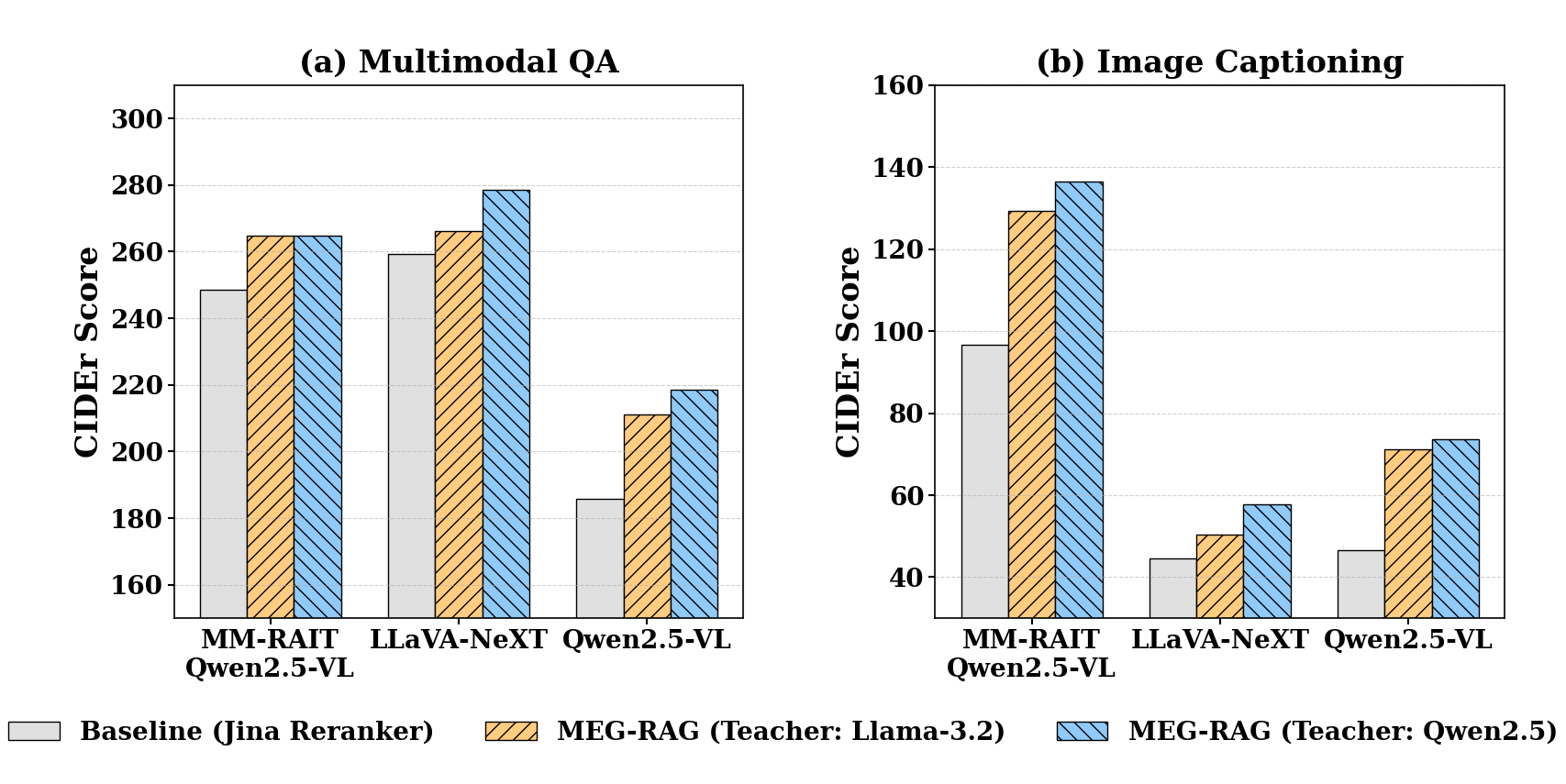}
    \vspace{-2em}
    \caption{Ablation study on Teacher Model Selection.}
    \label{fig:ablation_teacher}
\end{figure}

\textbf{Impact of Teacher Model Selection.} 
A critical concern in distillation-based approaches is whether the student model (reranker) learns generalizable information utility or merely mimics the biases of a specific teacher. To investigate this, we compared our default teacher (Qwen2.5-VL) with Llama-3.2-11B-Vision \cite{meta2024llama}, a model from a different architectural family. As illustrated in Figure \ref{fig:ablation_teacher}, the MEG-RAG is remarkably robust to the choice of teacher model. In all evaluated MLLMs and tasks, both teacher variants consistently and significantly outperform the baseline. For instance, on the MMQA task, the Llama-teacher variant (264.92 CIDEr) achieves performance almost identical to the Qwen-teacher variant (264.78 CIDEr), both surpassing the baseline by over 16 points.
This consistent improvement indicates that MEG primarily models evidence utility, rather than merely reflecting the bias of a specific teacher model.


\subsection{Sensitivity Analysis}

\begin{figure}[ht]
    \centering
    \includegraphics[width=\linewidth]{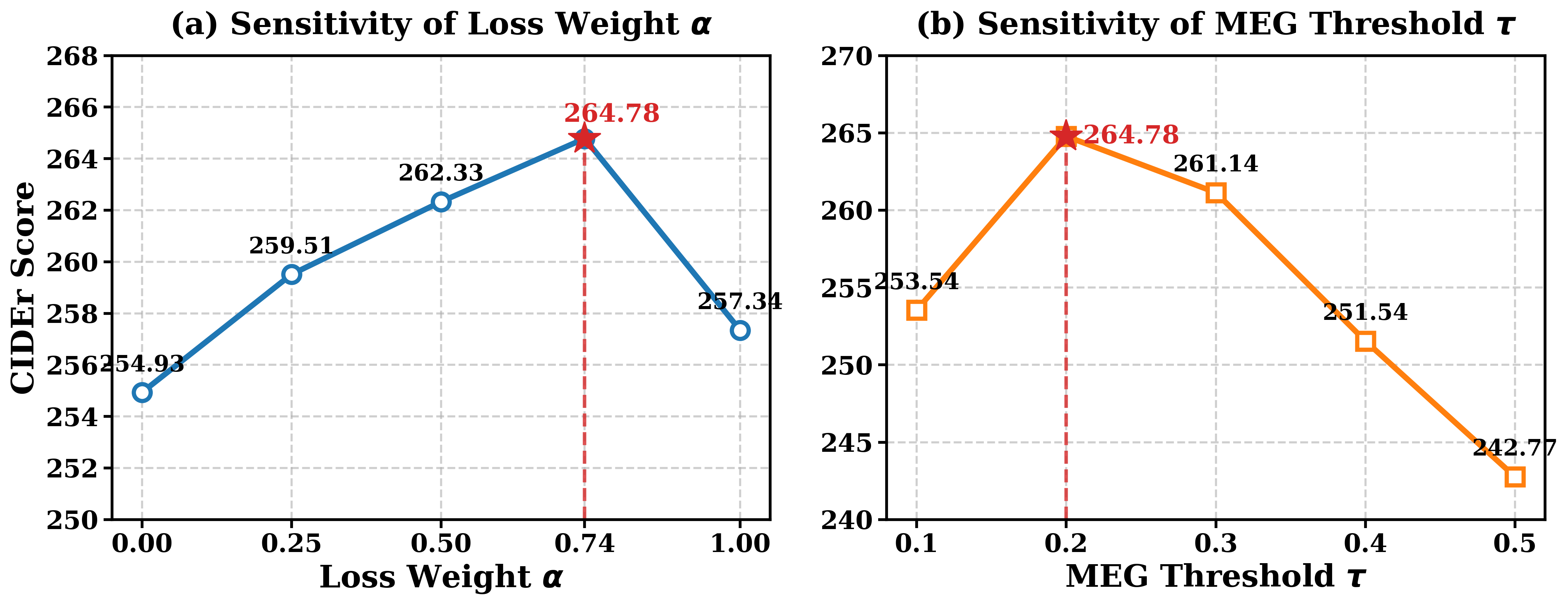} 
    \vspace{-1em}
    \caption{Sensitivity analysis of (a) the loss weight $\alpha$ and (b) the MEG labeling threshold $\tau$.}
    \label{fig:sensitivity}
\end{figure}

\textbf{Sensitivity of Loss Weight $\alpha$.} 
The hyperparameter $\alpha$ in Equation (\ref{eq:total_loss}) balances the CE loss and RankNet loss. To evaluate its impact, we conducted a grid search for $\alpha \in \{0.0, 0.25, 0.5, 0.74, 1.0\}$ and plotted the CIDEr performance of MEG-RAG (Top-3) with MM-RAIT-Qwen2.5-VL on the validation set. As illustrated in Figure \ref{fig:sensitivity}(a), relying solely on the RankNet loss ($\alpha=0.0$) or the CE loss ($\alpha=1.0$) yields sub-optimal results. The performance peaks around $\alpha=0.74$, demonstrating that a properly balanced hybrid loss is critical for optimizing the reranker.

\textbf{Sensitivity of MEG Threshold $\tau$.} 
We further analyzed the sensitivity of the positive/negative labeling threshold $\tau$ used during dataset construction. We tested varying the MEG boundary from 0.1 to 0.5 (i.e., assigning positive labels for $\mathrm{MEG} > \tau$ and negative for $\mathrm{MEG} < -\tau$). As shown in Figure \ref{fig:sensitivity}(b), a low threshold ($\tau=0.1$) introduces excessive noise from negligible documents, degrading performance. Conversely, an overly strict threshold ($\tau \ge 0.4$) severely limits the number of available training samples, leading to overfitting. The threshold of $\tau=0.2$ provides the optimal balance between dataset scale and signal purity.

\section{Conclusion}

In this paper, we present MEG-RAG, a unified framework for evidence selection in Multimodal Retrieval-Augmented Generation. MEG-RAG introduces Multi-modal Evidence Grounding (MEG), a principled metric that quantifies the contribution of retrieved multimodal evidence through a stabilized computation process. Based on MEG, we develop a hybrid-loss training strategy that enables a reranker to effectively distinguish and prioritize high-value multimodal documents. Experiments on the $M^2$RAG benchmark show that MEG-RAG consistently outperforms standard MRAG and existing reranking methods. Moreover, MEG-RAG is model-agnostic and generalizes well across various multimodal tasks, demonstrating its robustness and transferability in multimodal evidence selection.




\bibliographystyle{ACM-Reference-Format}
\bibliography{sample-base}


\begin{thebibliography}{20}


\ifx \showCODEN    \undefined \def \showCODEN     #1{\unskip}     \fi
\ifx \showISBNx    \undefined \def \showISBNx     #1{\unskip}     \fi
\ifx \showISBNxiii \undefined \def \showISBNxiii  #1{\unskip}     \fi
\ifx \showISSN     \undefined \def \showISSN      #1{\unskip}     \fi
\ifx \showLCCN     \undefined \def \showLCCN      #1{\unskip}     \fi
\ifx \shownote     \undefined \def \shownote      #1{#1}          \fi
\ifx \showarticletitle \undefined \def \showarticletitle #1{#1}   \fi
\ifx \showURL      \undefined \def \showURL       {\relax}        \fi
\providecommand\bibfield[2]{#2}
\providecommand\bibinfo[2]{#2}
\providecommand\natexlab[1]{#1}
\providecommand\showeprint[2][]{arXiv:#2}

\bibitem[Abootorabi et~al\mbox{.}(2025)]%
        {abootorabi2025ask}
\bibfield{author}{\bibinfo{person}{Mohammad~Mahdi Abootorabi}, \bibinfo{person}{Amirhosein Zobeiri}, \bibinfo{person}{Mahdi Dehghani}, \bibinfo{person}{Mohammadali Mohammadkhani}, \bibinfo{person}{Bardia Mohammadi}, \bibinfo{person}{Omid Ghahroodi}, \bibinfo{person}{Mahdieh~Soleymani Baghshah}, {and} \bibinfo{person}{Ehsaneddin Asgari}.} \bibinfo{year}{2025}\natexlab{}.
\newblock \showarticletitle{Ask in any modality: A comprehensive survey on multimodal retrieval-augmented generation}.
\newblock \bibinfo{journal}{\emph{arXiv preprint arXiv:2502.08826}} (\bibinfo{year}{2025}).
\newblock


\bibitem[AI(2024)]%
        {meta2024llama}
\bibfield{author}{\bibinfo{person}{Meta AI}.} \bibinfo{year}{2024}\natexlab{}.
\newblock \showarticletitle{Llama 3.2: Revolutionizing edge ai and vision with open, customizable models}.
\newblock \bibinfo{journal}{\emph{Meta AI Blog.}}  \bibinfo{volume}{20} (\bibinfo{year}{2024}), \bibinfo{pages}{2024}.
\newblock


\bibitem[Bai et~al\mbox{.}(2025)]%
        {bai2025qwen2}
\bibfield{author}{\bibinfo{person}{Shuai Bai}, \bibinfo{person}{Keqin Chen}, \bibinfo{person}{Xuejing Liu}, \bibinfo{person}{Jialin Wang}, \bibinfo{person}{Wenbin Ge}, \bibinfo{person}{Sibo Song}, \bibinfo{person}{Kai Dang}, \bibinfo{person}{Peng Wang}, \bibinfo{person}{Shijie Wang}, \bibinfo{person}{Jun Tang}, {et~al\mbox{.}}} \bibinfo{year}{2025}\natexlab{}.
\newblock \showarticletitle{Qwen2. 5-VL Technical Report}.
\newblock \bibinfo{journal}{\emph{eprint arXiv: 2502.13923}} (\bibinfo{year}{2025}).
\newblock


\bibitem[Burges et~al\mbox{.}(2005)]%
        {burges2005learning}
\bibfield{author}{\bibinfo{person}{Chris Burges}, \bibinfo{person}{Tal Shaked}, \bibinfo{person}{Erin Renshaw}, \bibinfo{person}{Ari Lazier}, \bibinfo{person}{Matt Deeds}, \bibinfo{person}{Nicole Hamilton}, {and} \bibinfo{person}{Greg Hullender}.} \bibinfo{year}{2005}\natexlab{}.
\newblock \showarticletitle{Learning to rank using gradient descent}. In \bibinfo{booktitle}{\emph{Proceedings of the 22nd international conference on Machine learning}}. \bibinfo{pages}{89--96}.
\newblock


\bibitem[Chen et~al\mbox{.}(2024)]%
        {chen2024mllm}
\bibfield{author}{\bibinfo{person}{Zhanpeng Chen}, \bibinfo{person}{Chengjin Xu}, \bibinfo{person}{Yiyan Qi}, {and} \bibinfo{person}{Jian Guo}.} \bibinfo{year}{2024}\natexlab{}.
\newblock \showarticletitle{{MLLM} Is a Strong Reranker: Advancing Multimodal Retrieval-augmented Generation via Knowledge-enhanced Reranking and Noise-injected Training}.
\newblock \bibinfo{journal}{\emph{CoRR}}  \bibinfo{volume}{abs/2407.21439} (\bibinfo{year}{2024}).
\newblock


\bibitem[Church and Hanks(1990)]%
        {church1990word}
\bibfield{author}{\bibinfo{person}{Kenneth Church} {and} \bibinfo{person}{Patrick Hanks}.} \bibinfo{year}{1990}\natexlab{}.
\newblock \showarticletitle{Word association norms, mutual information, and lexicography}.
\newblock \bibinfo{journal}{\emph{Computational linguistics}} \bibinfo{volume}{16}, \bibinfo{number}{1} (\bibinfo{year}{1990}), \bibinfo{pages}{22--29}.
\newblock


\bibitem[Huang et~al\mbox{.}(2025a)]%
        {huang2025towards}
\bibfield{author}{\bibinfo{person}{Chengkai Huang}, \bibinfo{person}{Junda Wu}, \bibinfo{person}{Yu Xia}, \bibinfo{person}{Zixu Yu}, \bibinfo{person}{Ruhan Wang}, \bibinfo{person}{Tong Yu}, \bibinfo{person}{Ruiyi Zhang}, \bibinfo{person}{Ryan~A Rossi}, \bibinfo{person}{Branislav Kveton}, \bibinfo{person}{Dongruo Zhou}, {et~al\mbox{.}}} \bibinfo{year}{2025}\natexlab{a}.
\newblock \showarticletitle{Towards agentic recommender systems in the era of multimodal large language models}.
\newblock \bibinfo{journal}{\emph{arXiv preprint arXiv:2503.16734}} (\bibinfo{year}{2025}).
\newblock


\bibitem[Huang et~al\mbox{.}(2025b)]%
        {huang2025embedding}
\bibfield{author}{\bibinfo{person}{Chengkai Huang}, \bibinfo{person}{Yu Xia}, \bibinfo{person}{Rui Wang}, \bibinfo{person}{Kaige Xie}, \bibinfo{person}{Tong Yu}, \bibinfo{person}{Julian McAuley}, {and} \bibinfo{person}{Lina Yao}.} \bibinfo{year}{2025}\natexlab{b}.
\newblock \showarticletitle{Embedding-informed adaptive retrieval-augmented generation of large language models}. In \bibinfo{booktitle}{\emph{Proceedings of the 31st International Conference on Computational Linguistics}}. \bibinfo{pages}{1403--1412}.
\newblock


\bibitem[Jiao et~al\mbox{.}(2026a)]%
        {jiao2026doctor}
\bibfield{author}{\bibinfo{person}{Shuguang Jiao}, \bibinfo{person}{Chengkai Huang}, \bibinfo{person}{Shuhan Qi}, \bibinfo{person}{Xuan Wang}, \bibinfo{person}{Yifan Li}, {and} \bibinfo{person}{Lina Yao}.} \bibinfo{year}{2026}\natexlab{a}.
\newblock \showarticletitle{Doctor-RAG: Failure-Aware Repair for Agentic Retrieval-Augmented Generation}.
\newblock \bibinfo{journal}{\emph{arXiv preprint arXiv:2604.00865}} (\bibinfo{year}{2026}).
\newblock


\bibitem[Jiao et~al\mbox{.}(2026b)]%
        {jiao2026prunerag}
\bibfield{author}{\bibinfo{person}{Shuguang Jiao}, \bibinfo{person}{Xinyu Xiao}, \bibinfo{person}{Yunfan Wei}, \bibinfo{person}{Shuhan Qi}, \bibinfo{person}{Chengkai Huang}, \bibinfo{person}{Quan~Z Sheng}, {and} \bibinfo{person}{Lina Yao}.} \bibinfo{year}{2026}\natexlab{b}.
\newblock \showarticletitle{PruneRAG: Confidence-Guided Query Decomposition Trees for Efficient Retrieval-Augmented Generation}. In \bibinfo{booktitle}{\emph{Proceedings of the ACM Web Conference 2026}}. \bibinfo{pages}{1923--1934}.
\newblock


\bibitem[{Jina AI}(2025)]%
        {jina-m0-blog}
\bibfield{author}{\bibinfo{person}{{Jina AI}}.} \bibinfo{year}{2025}\natexlab{}.
\newblock \bibinfo{title}{Jina Reranker M0: Multilingual \& Multimodal Document Reranker}.
\newblock


\bibitem[Kauf et~al\mbox{.}(2024)]%
        {kauf2024log}
\bibfield{author}{\bibinfo{person}{Carina Kauf}, \bibinfo{person}{Emmanuele Chersoni}, \bibinfo{person}{Alessandro Lenci}, \bibinfo{person}{Evelina Fedorenko}, {and} \bibinfo{person}{Anna~A Ivanova}.} \bibinfo{year}{2024}\natexlab{}.
\newblock \showarticletitle{Log probabilities are a reliable estimate of semantic plausibility in base and instruction-tuned language models}. In \bibinfo{booktitle}{\emph{Proceedings of the 7th BlackboxNLP Workshop: Analyzing and Interpreting Neural Networks for NLP}}. \bibinfo{pages}{263--277}.
\newblock


\bibitem[Li et~al\mbox{.}(2024)]%
        {li2024llava}
\bibfield{author}{\bibinfo{person}{Feng Li}, \bibinfo{person}{Renrui Zhang}, \bibinfo{person}{Hao Zhang}, \bibinfo{person}{Yuanhan Zhang}, \bibinfo{person}{Bo Li}, \bibinfo{person}{Wei Li}, \bibinfo{person}{Zejun Ma}, {and} \bibinfo{person}{Chunyuan Li}.} \bibinfo{year}{2024}\natexlab{}.
\newblock \showarticletitle{Llava-next-interleave: Tackling multi-image, video, and 3d in large multimodal models}.
\newblock \bibinfo{journal}{\emph{arXiv preprint arXiv:2407.07895}} (\bibinfo{year}{2024}).
\newblock


\bibitem[Liu et~al\mbox{.}(2025)]%
        {liu2025benchmarking}
\bibfield{author}{\bibinfo{person}{Zhenghao Liu}, \bibinfo{person}{Xingsheng Zhu}, \bibinfo{person}{Tianshuo Zhou}, \bibinfo{person}{Xinyi Zhang}, \bibinfo{person}{Xiaoyuan Yi}, \bibinfo{person}{Yukun Yan}, \bibinfo{person}{Ge Yu}, {and} \bibinfo{person}{Maosong Sun}.} \bibinfo{year}{2025}\natexlab{}.
\newblock \showarticletitle{Benchmarking retrieval-augmented generation in multi-modal contexts}. In \bibinfo{booktitle}{\emph{Proceedings of the 33rd ACM International Conference on Multimedia}}. \bibinfo{pages}{4817--4826}.
\newblock


\bibitem[Lou et~al\mbox{.}(2025)]%
        {lou2025speechagent}
\bibfield{author}{\bibinfo{person}{Haowei Lou}, \bibinfo{person}{Chengkai Huang}, \bibinfo{person}{Hye-young Paik}, \bibinfo{person}{Yongquan Hu}, \bibinfo{person}{Aaron Quigley}, \bibinfo{person}{Wen Hu}, {and} \bibinfo{person}{Lina Yao}.} \bibinfo{year}{2025}\natexlab{}.
\newblock \showarticletitle{SpeechAgent: An End-to-End Mobile Infrastructure for Speech Impairment Assistance}.
\newblock \bibinfo{journal}{\emph{arXiv preprint arXiv:2510.20113}} (\bibinfo{year}{2025}).
\newblock


\bibitem[Mei et~al\mbox{.}(2025)]%
        {mei2025survey}
\bibfield{author}{\bibinfo{person}{Lang Mei}, \bibinfo{person}{Siyu Mo}, \bibinfo{person}{Zhihan Yang}, {and} \bibinfo{person}{Chong Chen}.} \bibinfo{year}{2025}\natexlab{}.
\newblock \showarticletitle{A survey of multimodal retrieval-augmented generation}.
\newblock \bibinfo{journal}{\emph{arXiv preprint arXiv:2504.08748}} (\bibinfo{year}{2025}).
\newblock


\bibitem[Mortaheb et~al\mbox{.}(2025)]%
        {mortaheb2025re}
\bibfield{author}{\bibinfo{person}{Matin Mortaheb}, \bibinfo{person}{Mohammad A~Amir Khojastepour}, \bibinfo{person}{Srimat~T Chakradhar}, {and} \bibinfo{person}{Sennur Ulukus}.} \bibinfo{year}{2025}\natexlab{}.
\newblock \showarticletitle{Re-ranking the context for multimodal retrieval augmented generation}.
\newblock \bibinfo{journal}{\emph{arXiv preprint arXiv:2501.04695}} (\bibinfo{year}{2025}).
\newblock


\bibitem[Wang et~al\mbox{.}(2026)]%
        {wang2026scenealign}
\bibfield{author}{\bibinfo{person}{Chuhan Wang}, \bibinfo{person}{Xintong Li}, \bibinfo{person}{Jennifer~Yuntong Zhang}, \bibinfo{person}{Junda Wu}, \bibinfo{person}{Chengkai Huang}, \bibinfo{person}{Lina Yao}, \bibinfo{person}{Julian McAuley}, {and} \bibinfo{person}{Jingbo Shang}.} \bibinfo{year}{2026}\natexlab{}.
\newblock \showarticletitle{SceneAlign: Aligning Multimodal Reasoning to Scene Graphs in Complex Visual Scenes}.
\newblock \bibinfo{journal}{\emph{arXiv preprint arXiv:2601.05600}} (\bibinfo{year}{2026}).
\newblock


\bibitem[Xu et~al\mbox{.}(2025)]%
        {xu2025mm}
\bibfield{author}{\bibinfo{person}{Mingjun Xu}, \bibinfo{person}{Jinhan Dong}, \bibinfo{person}{Jue Hou}, \bibinfo{person}{Zehui Wang}, \bibinfo{person}{Sihang Li}, \bibinfo{person}{Zhifeng Gao}, \bibinfo{person}{Renxin Zhong}, {and} \bibinfo{person}{Hengxing Cai}.} \bibinfo{year}{2025}\natexlab{}.
\newblock \showarticletitle{MM-R5: MultiModal Reasoning-Enhanced ReRanker via Reinforcement Learning for Document Retrieval}.
\newblock \bibinfo{journal}{\emph{arXiv preprint arXiv:2506.12364}} (\bibinfo{year}{2025}).
\newblock


\bibitem[Ye et~al\mbox{.}(2026)]%
        {ye2026memweaver}
\bibfield{author}{\bibinfo{person}{Juexiang Ye}, \bibinfo{person}{Xue Li}, \bibinfo{person}{Xinyu Yang}, \bibinfo{person}{Chengkai Huang}, \bibinfo{person}{Lanshun Nie}, \bibinfo{person}{Lina Yao}, {and} \bibinfo{person}{Dechen Zhan}.} \bibinfo{year}{2026}\natexlab{}.
\newblock \showarticletitle{MemWeaver: Weaving Hybrid Memories for Traceable Long-Horizon Agentic Reasoning}.
\newblock \bibinfo{journal}{\emph{arXiv preprint arXiv:2601.18204}} (\bibinfo{year}{2026}).
\newblock


\end{thebibliography}

\appendix

\end{document}